\documentclass[11pt,a4paper]{article}
\usepackage[hyperref]{eacl2021}
\usepackage{times}
\usepackage{latexsym}

\usepackage{url}
\usepackage{latexsym}
\usepackage{soul}
\usepackage[utf8]{inputenc}
\usepackage{caption}
\usepackage{subcaption}
\usepackage{graphicx}
\graphicspath{{./figs/}}
\usepackage{amsmath}
\usepackage{amsthm}
\usepackage{booktabs}
\usepackage{algorithm}
\usepackage{algpseudocode}
\usepackage{mathtools}
\usepackage{multirow}
\usepackage{array}
\usepackage{tabularx}

\usepackage{microtype}

\aclfinalcopy 

\title{Do Multi-Hop Question Answering Systems Know \\ How to Answer the Single-Hop Sub-Questions?}

\author{Yixuan Tang \qquad Hwee Tou Ng \qquad  Anthony K.H. Tung\\ \\
  Department of Computer Science \\
  National University of Singapore \\
  \texttt{\{yixuan, nght, atung\}@comp.nus.edu.sg} \\}

\date{}

\begin{document}
	\newcolumntype{L}[1]{>{\raggedright\arraybackslash}m{#1}}
	\newcolumntype{C}[1]{>{\centering\arraybackslash}m{#1}}
	\newcolumntype{R}[1]{>{\raggedleft\arraybackslash}m{#1}}

\maketitle
\begin{abstract}
	Multi-hop question answering (QA) requires a model to retrieve and integrate information from multiple passages to answer a question. Rapid progress has been made on multi-hop QA systems with regard to standard evaluation metrics, including EM and F1. However, by simply evaluating the correctness of the answers, it is unclear to what extent these systems have learned the ability to perform multi-hop reasoning. In this paper, we propose an additional sub-question evaluation for the multi-hop QA dataset HotpotQA, in order to shed some light on explaining the reasoning process of QA systems in answering complex questions. We adopt a neural decomposition model to generate sub-questions for a multi-hop question, followed by extracting the corresponding sub-answers. Contrary to our expectation, multiple state-of-the-art multi-hop QA models fail to answer a large portion of sub-questions, although the corresponding multi-hop questions are correctly answered. Our work takes a step forward towards building a more explainable multi-hop QA system.
\end{abstract}

\section{Introduction}
\label{intro}

\begin{figure*}[t!]
	\centering
	\includegraphics[width=\textwidth]{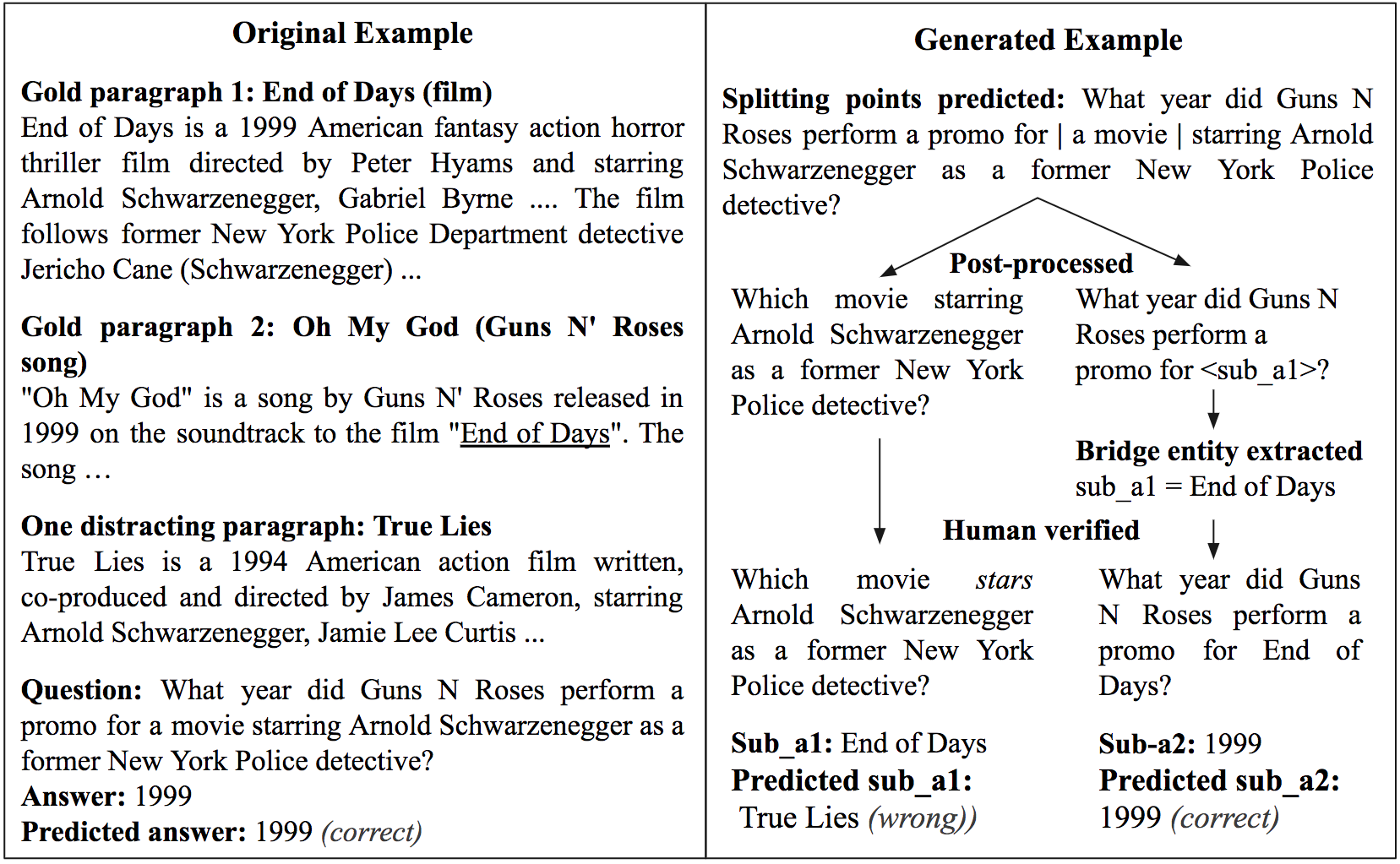}
	\caption{An illustrating example from the HotpotQA dataset in the distractor setting, with our construction procedure to generate an evaluation example. We only show one out of eight distracting paragraphs provided in the context due to paper length constraint.}
	\label{fig1:eg1}
\end{figure*}

Rapid progress has been made in the field of question answering (QA), thanks to the release of many large-scale, high-quality QA datasets. Early datasets \cite{DBLP:conf/nips/HermannKGEKSB15,DBLP:conf/emnlp/RajpurkarZLL16,DBLP:conf/acl/RajpurkarJL18,DBLP:conf/rep4nlp/TrischlerWYHSBS17, DBLP:conf/acl/JoshiCWZ17} mainly consist of single-hop questions, where an answer with supporting justification can be found within a short segment of text. These benchmarks focus on evaluating QA models' ability to perform local pattern matching between a passage and a question. Existing models \cite{DBLP:journals/corr/abs-1909-11942,DBLP:journals/corr/abs-1908-05147} have achieved super-human performance. Recently, multi-hop QA datasets \cite{DBLP:conf/naacl/KhashabiCRUR18, journals/tacl/WelblSR18, DBLP:conf/emnlp/Yang0ZBCSM18} have gained increasing attention. They require models to retrieve multiple pieces of supporting evidence from different documents and to reason over the evidence collected to answer a question. The standard evaluation metrics of QA datasets include exact match (EM) and F1 scores averaged over the test set. HotpotQA \cite{DBLP:conf/emnlp/Yang0ZBCSM18} also provides sentence-level supporting facts required for reasoning. However, providing supporting sentences is not sufficient for us to interpret the choice of an answer for end-to-end complex QA systems. It is unclear whether the systems have performed the desired multi-hop reasoning to reach the correct answer.

In this work, we propose an additional evaluation scheme to test multi-hop QA systems' performance on answering the single-hop sub-questions of a multi-hop question. When designing a multi-hop question, we expect it to require QA models to retrieve a chain of sentences as evidence and then reasoning over them to answer the question. Evaluating QA models on sub-questions helps us to understand their behavior on each hop of the reasoning process. In addition, it evaluates whether multi-hop QA models can generalize well on simpler questions. Figure 1 presents an illustrating example. A successful complex QA model should be able to answer the two sub-questions ``\textit{Which movie stars Arnold Schwarzenegger as a former New York Police detective}" and ``\textit{What year did Guns N Roses perform a promo for End of Days}" if it understands the underlying reasoning process for the original multi-hop question. 

We focus on the HotpotQA \cite{DBLP:conf/emnlp/Yang0ZBCSM18} dataset under the distractor setting, in which multi-hop questions are asked over several Wikipedia paragraphs. We create the evaluation dataset by generating the sub-questions and then extracting their answers automatically. The candidate sub-questions and intermediate answers are then manually verified, which results in 1,000 sub-question evaluation examples. It is surprising to find that all three top-performing models which we experiment with fail to answer a large portion of sub-questions (49.8\% to 60.4\%), although their corresponding multi-hop questions are correctly answered.

Previous work has investigated the necessity of multi-hop reasoning on HotpotQA dataset. \newcite{DBLP:conf/acl/JiangB19} construct distracting paragraphs adversarially to demonstrate that models learn to exploit reasoning shortcuts to locate the answer rather than performing multi-hop reasoning. \newcite{DBLP:conf/naacl/ChenD19} show that a sentence-factored model can solve a large number of questions in HotpotQA, suggesting multi-hop reasoning is not really needed. \newcite{DBLP:conf/acl/MinWSGHZ19} also achieve similar result using a single-hop BERT-based model. Our sub-question evaluation is complementary to these approaches. While existing work shows the lack of multi-hop reasoning by limiting or adding text input to QA models, we provide sub-questions and intermediate answers explicitly to interpret model behavior on each hop of the reasoning process. It can be used as a complementary metric to ensure that models which can correctly answer both intermediate sub-questions and the final multi-hop question actually go through the reasoning steps as desired. Our work takes a step forward towards building a more explainable multi-hop QA system.

\section{Construction of Evaluation Examples}

In this section, we introduce our semi-automatic approach to generate two sub-questions and their corresponding answers for multi-hop questions from the HotpotQA dataset. As shown in Figure \ref{fig1:eg1}, the evaluation examples are generated in three steps. First, we decompose each source question into several sub-strings by predicting the breaking points and post-process them to generate two sub-questions. Then, the answers for the sub-questions are extracted from the paragraphs using some heuristics. Lastly, the candidate evaluation examples generated are sent for human verification. We first introduce the HotpotQA dataset and then elaborate on each step of the construction pipeline.

\subsection{HotpotQA}
HotpotQA contains 113K crowd-sourced multi-hop QA pairs on Wikipedia articles. We focus on bridge-type questions that actually require multiple steps of reasoning under the distractor setting. During the construction of such an example in HotpotQA, two related paragraphs $p_{gold1}, p_{gold2}$ from different Wikipedia articles titled $t_{gold1}, t_{gold2}$ are presented to crowd-workers. The two paragraphs are related since the text content in one paragraph contains the title entity of the other paragraph. This shared title entity is referred to as the bridge entity. Using Figure \ref{fig1:eg1} as an example, the second paragraph about \textit{Oh My God} contains the title entity of the first paragraph, \textit{End of Days} (underlined). Thus, \textit{End of Days} is referred as the bridge entity. The crowd-workers are encouraged to ask a multi-hop question using both paragraphs and to annotate the supporting sentences which help to determine the answer. Then, eight other related distracting paragraphs are retrieved from Wikipedia and mixed with the two gold paragraphs to serve as the context for the question. Given an example $E = \{C, q, a\}$ from HotpotQA, we aim to generate an evaluation example $E' = \{C, q, a, sub\_q_1, sub\_a_1, sub\_q_2, sub\_a_2\}$, where $sub\_q_1$ and $sub\_q_2$ are the two sub-questions, and $sub\_a_1$ and $sub\_a_2$ are their corresponding answers.

\subsection{Sub-Question Generation}

Given a multi-hop question, the first step is to decompose it into sub-questions. We adopt the model introduced in DecompRC \cite{DBLP:conf/acl/MinZZH19} to generate the sub-questions using a copying and editing mechanism. The multi-hop question is first converted into BERT word embeddings \cite{DBLP:conf/naacl/DevlinCLT19}, and then sent to a fully connected neural network to predict the splitting points. It is trained on 400 annotated examples. The separated text spans are post-processed to form the two sub-questions, following a set of handcrafted rules.

\subsection{Intermediate Answer Extraction}

One particular characteristic of bridge-type questions from HotpotQA is that the two gold paragraphs are linked by a bridge entity. Since the crowd-workers are required to form a multi-hop question which makes use of information from both paragraphs, there is a high probability that the bridge entity is the answer to the first sub-question. For the example shown in Figure \ref{fig1:eg1}, \textit{End of Days} in gold paragraph 2 is the bridge entity. It is also the intermediate answer for the multi-hop question, i.e., the answer for the first sub-question.

Three different situations are considered in order to extract the bridge entity. First, if the title entity $E_A$ of paragraph $A$ occurs in the other paragraph $B$, while the title entity $E_B$ of $B$ does not occur in $A$, then $E_A$ is recognized as the bridge entity. Second, if neither $E_A$ nor $E_B$ is contained in the other paragraph, then the title entity with more overlapping text with the other paragraph is chosen as the bridge entity (since sometimes the alias of the Wikipedia title is used in the paragraph). Lastly, if both $E_A$ and $E_B$ appear in the other paragraph, then the title entity which does not appear in both the question and the answer is chosen as the bridge entity, since an entity mentioned in the multi-hop question or included in the final answer is unlikely to be the bridge entity. The bridge entity is set to be unidentified if none or both of the title entities satisfy at least one of the requirements. As illustrated in Figure \ref{fig1:eg1}, once the bridge entity is retrieved, the blank in the second sub-question will be updated. The answer to the second sub-question should be the same as the original multi-hop question.

 \begin{table}[t]
	\centering
	\begin{tabular}{C{0.5cm} L{3.2cm} L{2.8cm}}
		\hline
		Case & Gold Answer & Predicted Answer \\ \hline
		1& from 1986 to 2013 & 1986 to 2013 \\ \hline
		2& City of Angles (film) & City of Angles \\ \hline
		3& Mondelez International, Inc.	&  the company Mondelez International\\ \hline	
	\end{tabular}
	\caption{Examples of partially matched answer string pairs.}	
	\label{tab1:partial_match}
\end{table}

\begin{table}[t]
	\centering
	\begin{tabular}{C{1.5cm} | L{0.4cm} C{0.65cm} | C{0.4cm} C{0.65cm} | C{0.5cm} C{0.65cm}} 
		\hline
		\multirow{2}{*}{Model} & \multicolumn{2}{c|}{$q$} & \multicolumn{2}{c|}{$q_{sub1}$} & \multicolumn{2}{c}{ $q_{sub2}$}    	\\ \cline{2-7}
		& EM  & F1 & EM  & F1 & EM  & F1 \\ \hline
		DFGN & {\fontsize{10}{12}\selectfont  58.1 }
		& {\fontsize{10}{12}\selectfont 71.96 	}
		& {\fontsize{10}{12}\selectfont  54.6 }
		& {\fontsize{10}{12}\selectfont 68.54 }
		& {\fontsize{10}{12}\selectfont 49.3 }
		&{\fontsize{10}{12}\selectfont  60.83} \\ \hline
		
		{\fontsize{10}{12}\selectfont DecompRC} 
		&    {\fontsize{10}{12}\selectfont 63.1 }
		&  {\fontsize{10}{12}\selectfont 77.61}
		&  {\fontsize{10}{12}\selectfont  61 }
		&  {\fontsize{10}{12}\selectfont 75.21 }
		&  {\fontsize{10}{12}\selectfont 56.8 }
		&  {\fontsize{10}{12}\selectfont 70.77}\\ \hline
		
		CogQA 
		&  {\fontsize{10}{12}\selectfont  53.2}
		 &  {\fontsize{10}{12}\selectfont 67.82}
		  &  {\fontsize{10}{12}\selectfont 58.6 }
		  &  {\fontsize{10}{12}\selectfont 69.65 }
		  &   {\fontsize{10}{12}\selectfont 54} 
		  &  {\fontsize{10}{12}\selectfont 68.49}\\ \hline
	\end{tabular}
	\caption{EM and F1 scores of models on 1,000 human-verified sub-question evaluation examples.}
	\label{tab2:emf1}
\end{table}

\subsection{Human Verification}

Sub-question generation and intermediate answer extraction help to efficiently generate candidate sub-questions and their answers. To ensure the quality of the evaluation dataset, the examples generated are manually verified. For each example, we present to an annotator the original multi-hop question, the answer, two sub-questions generated and their answers, and two gold paragraphs. Questions that actually do not require multi-hop reasoning or with the wrong answer (due to wrong annotation by the HotpotQA crowd workers) are first filtered out. Then, the annotator is required to review whether $sub\_q_1$ and $sub\_q_2$ are two syntactically and semantically correct sub-questions of $q$ and whether $sub\_a_1$ and $sub\_a_2$ are valid and to correct them if not. In total, a sample of 1,000 examples generated for the HotpotQA development set are manually verified for use in our evaluation\footnote{The verified dataset is available at \url{https://github.com/yxxytang/subqa}}.

\section{Experiments and Results}

 In order to interpret the behavior of existing models on each hop of the reasoning process required for multi-hop questions and to determine their ability to answer simple questions, we perform sub-question evaluation on three published top-performing QA models with publicly available open-source code: DFGN \cite{DBLP:conf/acl/QiuXQZLZY19}, DecompRC \cite{DBLP:conf/acl/MinZZH19}, and CogQA \cite{DBLP:conf/acl/DingZCYT19}. For all experiments, we measure EM and F1 scores for $q$, $sub\_q_1$, and $sub\_q_2$ on 1,000 human-verified examples. To measure the correctness of a predicted answer, we first use exact string match as the only metric. However, during error analysis, we find that many predicted answers that partially match the gold answers should also be regarded as correct. Some representative examples are shown in Table \ref{tab1:partial_match}. Although these predicted answers have zero EM scores, they are semantically equivalent to the correct answers given. Therefore, we define a more flexible metric named partial match (PM) as an additional evaluation of correctness. Given a gold answer text span $a_g$ and a predicted answer text $a_p$, they partially match if either one of the following two requirements is satisfied:

\begin{align}
	\begin{aligned}
		f1& > 0.8  \\
		f1 > 0.6 \land \{(a_g \: contain&s \: a_p) \lor (a_p \: contains \: a_g)\}
	\end{aligned}
\end{align}

\begin{table}[t]
	\centering	
	\begin{tabular}{C{0.4cm} C{0.5cm} C{0.6cm} | C{0.7cm} C{1.6cm} C{1.0cm}}\hline
    	$q$ & $q_{sub1}$ & $q_{sub2}$  & DFGN & DecompRC & CogQA \\  \hline
	    c   & c  & c  & 23.0   & 31.3     & 26.7          \\ \hline
		c   & c & w    & 9.7  & 7.2      & 5.8         \\ \hline
		c   & w   & c  & 17.9 & 19.1     & 17.8            \\ \hline
		c  & w   & w    & 7.5  & 5.5      & 2.9              \\ \hline
		w  & c  & c  & 4.9  & 3.0        & 3.6           \\ \hline
		w    & c  & w  & 17.0   & 18.6     & 22.5        \\ \hline
		w   & w   & c  & 3.5  & 3.4      & 5.9           \\ \hline
		w    & w   & w    & 16.5 & 11.9     & 14.8          \\ \hline
	\end{tabular}
	\caption{Categorical EM statistics (\%) of sub-question evaluation for the three models. Under the first three columns, \textit{c} stands for \textit{correct} and \textit{w} stands for \textit{wrong}. For example, the second row shows the percentage of questions where models correctly answer both multi-hop question and the first sub-question but wrongly answer the second sub-question.}	
	\label{tab3:result}
\end{table}

\begin{table}[t]
	\centering	
	\begin{tabular}{C{0.4cm} C{0.5cm} C{0.6cm} | C{0.7cm} C{1.6cm}  C{1.0cm}}
		\hline
    	$q$ & $q_{sub1}$ & $q_{sub2}$    & DFGN & DecompRC & CogQA \\  \hline
	    c   & c  & c  &    36.3 & 47.4     & 40.9           \\ \hline
		c   & c & w    &   11.9 & 8.5      & 6.1      \\ \hline
		c   & w   & c  &   16.4 & 17.2     & 16.5         \\ \hline
		c  & w   & w    &  6.5  & 3.9      & 3.4           \\ \hline
		w  & c  & c  &      4.2  & 4.0        & 4.5      \\ \hline
		w    & c  & w  &   12.1 & 11.1     & 15.2        \\ \hline
		w   & w   & c  &     3.1  & 1.9      & 5.6        \\ \hline
		w    & w   & w  &     9.5  & 6.0    & 7.8 \\ \hline
	\end{tabular}
	\caption{Categorical PM statistics (\%) of sub-question evaluation for the three models.}	
	\label{tab4:result}
\end{table}

\begin{figure}[t]
	\centering
		\includegraphics[width=0.96\linewidth]{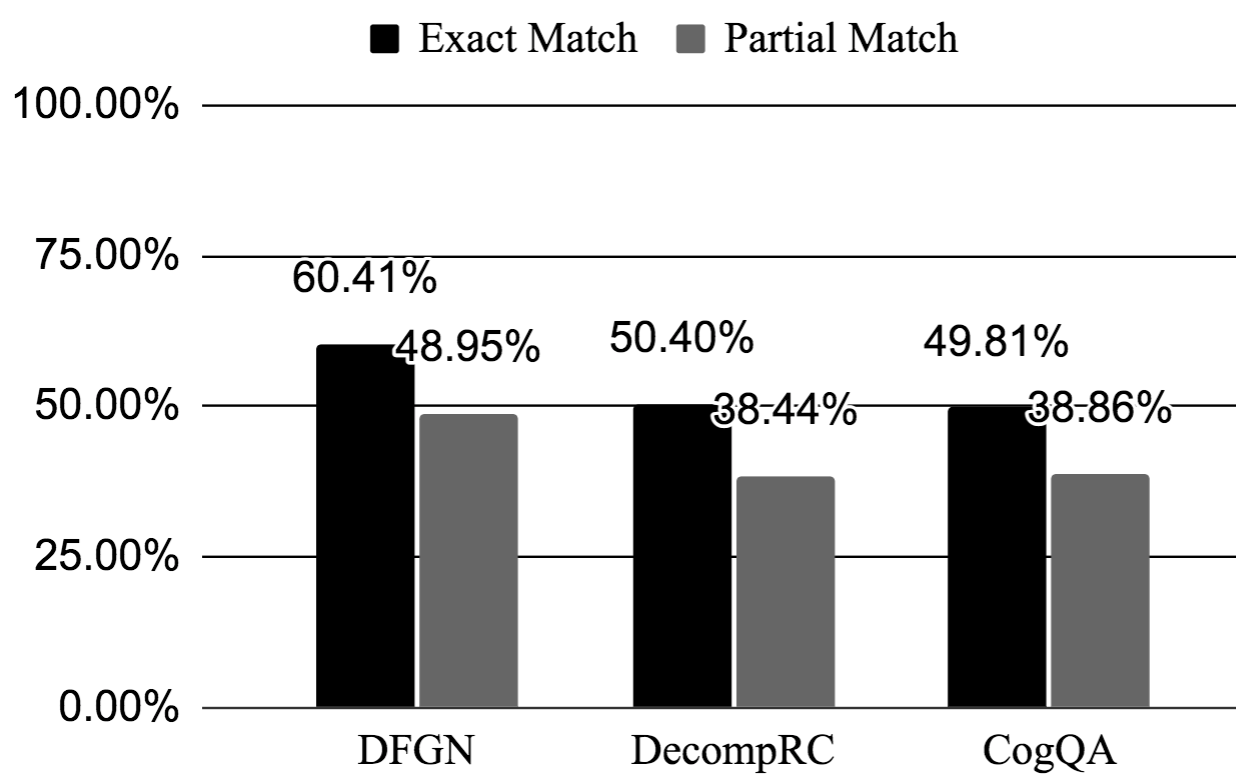}
		\caption{Model failure rates under EM and PM.}
		\label{fig:chart}
\end{figure}

 Table \ref{tab2:emf1} shows the performance of the three models on multi-hop questions and their single-hop sub-questions. Compared to multi-hop questions, the performance of DFGN and DecompRC drops on simpler sub-questions, especially on the second sub-questions (11.13 F1 reduction for DFGN and 6.84 F1 reduction for DecompRC). CogQA achieves slightly better performance on sub-questions, which shows that it is also able to answer single-hop questions. The EM and F1 scores are averaged over all examples. In order to understand whether models are able to answer the sub-questions of correctly answered multi-hop questions, we collect the correctness statistics with regard to each individual example. Table \ref{tab3:result} and Table \ref{tab4:result} present the results. The first four rows show the percentage of examples whose multi-hop question can be correctly answered. Among these examples, we notice that there is a high probability that the models fail to answer at least one of the sub-questions, as shown in rows 2 to 4. We refer to these examples as model failure cases. The percentage of model failure cases over all correctly answered multi-hop questions is defined as model failure rate. As shown in Figure \ref{fig:chart}, all three models evaluated have a high model failure rate, indicating that the models learn to answer the complex questions without exploring the multiple steps of reasoning process as desired. The same phenomenon appears when evaluated using exact match and the less strict partial match scores.

After analyzing the model failure cases, we observe a common phenomenon that there is a high similarity between the words in the second sub-question and the words near the answer in the context. The model has learned to answer multi-hop question by local pattern matching, instead of going through the multiple reasoning steps. For the example presented in Figure \ref{fig1:eg1}, the model may locate the answer ``\textit{1999}" for the multi-hop question by matching the surrounding words ``\textit{ Guns N Roses}" in the second sub-question. Despite answering the multi-hop question correctly, the model fails to identify the answer of the first sub-question which it is expected to retrieve as a multi-hop QA system.

\section{Conclusion}

We propose a new way to interpret whether multi-hop QA systems explore the multiple steps of reasoning over the evidence as desired by asking sub-questions. An automatic approach is designed to generate sub-questions for a multi-hop question. On a human-verified test set, our experiments demonstrate that top-performing multi-hop QA models fail to answer a large portion of sub-questions whose parent multi-hop questions can be correctly answered. We believe that progress on building complex QA systems that truly understand multi-hop reasoning is only possible if the evaluation metrics reward this kind of behavior. As an initial step towards a more explainable QA system, we hope our work would motivate the construction of multi-hop QA datasets with explicit reasoning paths annotated and the development of better multi-hop QA models.

\bibliography{eacl2021}

\begin{thebibliography}{17}
\expandafter\ifx\csname natexlab\endcsname\relax\def\natexlab#1{#1}\fi

\bibitem[{Chen and Durrett(2019)}]{DBLP:conf/naacl/ChenD19}
Jifan Chen and Greg Durrett. 2019.
\newblock Understanding dataset design choices for multi-hop reasoning.
\newblock In \emph{{NAACL-HLT}}, pages 4026--4032.

\bibitem[{Devlin et~al.(2019)Devlin, Chang, Lee, and
  Toutanova}]{DBLP:conf/naacl/DevlinCLT19}
Jacob Devlin, Ming{-}Wei Chang, Kenton Lee, and Kristina Toutanova. 2019.
\newblock {BERT:} pre-training of deep bidirectional transformers for language
  understanding.
\newblock In \emph{{NAACL-HLT}}, pages 4171--4186.

\bibitem[{Ding et~al.(2019)Ding, Zhou, Chen, Yang, and
  Tang}]{DBLP:conf/acl/DingZCYT19}
Ming Ding, Chang Zhou, Qibin Chen, Hongxia Yang, and Jie Tang. 2019.
\newblock Cognitive graph for multi-hop reading comprehension at scale.
\newblock In \emph{{ACL}}, pages 2694--2703.

\bibitem[{Hermann et~al.(2015)Hermann, Kocisk{\'{y}}, Grefenstette, Espeholt,
  Kay, Suleyman, and Blunsom}]{DBLP:conf/nips/HermannKGEKSB15}
Karl~Moritz Hermann, Tom{\'{a}}s Kocisk{\'{y}}, Edward Grefenstette, Lasse
  Espeholt, Will Kay, Mustafa Suleyman, and Phil Blunsom. 2015.
\newblock Teaching machines to read and comprehend.
\newblock In \emph{{NIPS}}, pages 1693--1701.

\bibitem[{Jiang and Bansal(2019)}]{DBLP:conf/acl/JiangB19}
Yichen Jiang and Mohit Bansal. 2019.
\newblock Avoiding reasoning shortcuts: Adversarial evaluation, training, and
  model development for multi-hop {QA}.
\newblock In \emph{{ACL}}, pages 2726--2736.

\bibitem[{Joshi et~al.(2017)Joshi, Choi, Weld, and
  Zettlemoyer}]{DBLP:conf/acl/JoshiCWZ17}
Mandar Joshi, Eunsol Choi, Daniel~S. Weld, and Luke Zettlemoyer. 2017.
\newblock Trivia{QA}: {A} large scale distantly supervised challenge dataset
  for reading comprehension.
\newblock In \emph{{ACL}}, pages 1601--1611.

\bibitem[{Khashabi et~al.(2018)Khashabi, Chaturvedi, Roth, Upadhyay, and
  Roth}]{DBLP:conf/naacl/KhashabiCRUR18}
Daniel Khashabi, Snigdha Chaturvedi, Michael Roth, Shyam Upadhyay, and Dan
  Roth. 2018.
\newblock Looking beyond the surface: {A} challenge set for reading
  comprehension over multiple sentences.
\newblock In \emph{{NAACL-HLT}}, pages 252--262.

\bibitem[{Lan et~al.(2020)Lan, Chen, Goodman, Gimpel, Sharma, and
  Soricut}]{DBLP:journals/corr/abs-1909-11942}
Zhenzhong Lan, Mingda Chen, Sebastian Goodman, Kevin Gimpel, Piyush Sharma, and
  Radu Soricut. 2020.
\newblock {ALBERT:} {A} lite {BERT} for self-supervised learning of language
  representations.
\newblock In \emph{{ICLR}}.

\bibitem[{Min et~al.(2019{\natexlab{a}})Min, Wallace, Singh, Gardner,
  Hajishirzi, and Zettlemoyer}]{DBLP:conf/acl/MinWSGHZ19}
Sewon Min, Eric Wallace, Sameer Singh, Matt Gardner, Hannaneh Hajishirzi, and
  Luke Zettlemoyer. 2019{\natexlab{a}}.
\newblock Compositional questions do not necessitate multi-hop reasoning.
\newblock In \emph{{ACL}}, pages 4249--4257.

\bibitem[{Min et~al.(2019{\natexlab{b}})Min, Zhong, Zettlemoyer, and
  Hajishirzi}]{DBLP:conf/acl/MinZZH19}
Sewon Min, Victor Zhong, Luke Zettlemoyer, and Hannaneh Hajishirzi.
  2019{\natexlab{b}}.
\newblock Multi-hop reading comprehension through question decomposition and
  rescoring.
\newblock In \emph{{ACL}}, pages 6097--6109.

\bibitem[{Qiu et~al.(2019)Qiu, Xiao, Qu, Zhou, Li, Zhang, and
  Yu}]{DBLP:conf/acl/QiuXQZLZY19}
Lin Qiu, Yunxuan Xiao, Yanru Qu, Hao Zhou, Lei Li, Weinan Zhang, and Yong Yu.
  2019.
\newblock Dynamically fused graph network for multi-hop reasoning.
\newblock In \emph{{ACL}}, pages 6140--6150.

\bibitem[{Rajpurkar et~al.(2018)Rajpurkar, Jia, and
  Liang}]{DBLP:conf/acl/RajpurkarJL18}
Pranav Rajpurkar, Robin Jia, and Percy Liang. 2018.
\newblock Know what you don't know: Unanswerable questions for {SQ}u{AD}.
\newblock In \emph{{ACL}}, pages 784--789.

\bibitem[{Rajpurkar et~al.(2016)Rajpurkar, Zhang, Lopyrev, and
  Liang}]{DBLP:conf/emnlp/RajpurkarZLL16}
Pranav Rajpurkar, Jian Zhang, Konstantin Lopyrev, and Percy Liang. 2016.
\newblock {SQ}u{AD}: 100, 000+ questions for machine comprehension of text.
\newblock In \emph{{EMNLP}}, pages 2383--2392.

\bibitem[{Trischler et~al.(2017)Trischler, Wang, Yuan, Harris, Sordoni,
  Bachman, and Suleman}]{DBLP:conf/rep4nlp/TrischlerWYHSBS17}
Adam Trischler, Tong Wang, Xingdi Yuan, Justin Harris, Alessandro Sordoni,
  Philip Bachman, and Kaheer Suleman. 2017.
\newblock News{QA}: {A} machine comprehension dataset.
\newblock In \emph{Rep4NLP@ACL}, pages 191--200.

\bibitem[{Welbl et~al.(2018)Welbl, Stenetorp, and
  Riedel}]{journals/tacl/WelblSR18}
Johannes Welbl, Pontus Stenetorp, and Sebastian Riedel. 2018.
\newblock Constructing datasets for multi-hop reading comprehension across
  documents.
\newblock In \emph{TACL}, pages 287--302.

\bibitem[{Yang et~al.(2018)Yang, Qi, Zhang, Bengio, Cohen, Salakhutdinov, and
  Manning}]{DBLP:conf/emnlp/Yang0ZBCSM18}
Zhilin Yang, Peng Qi, Saizheng Zhang, Yoshua Bengio, William~W. Cohen, Ruslan
  Salakhutdinov, and Christopher~D. Manning. 2018.
\newblock Hotpot{QA}: {A} dataset for diverse, explainable multi-hop question
  answering.
\newblock In \emph{{EMNLP}}, pages 2369--2380.

\bibitem[{Zhang et~al.(2020)Zhang, Wu, Zhou, Duan, Zhao, and
  Wang}]{DBLP:journals/corr/abs-1908-05147}
Zhuosheng Zhang, Yuwei Wu, Junru Zhou, Sufeng Duan, Hai Zhao, and Rui Wang.
  2020.
\newblock {SG-Net:} syntax-guided machine reading comprehension.
\newblock In \emph{{AAAI}}, pages 9636--9643.

\end{thebibliography}
\bibliographystyle{acl_natbib}

\end{document}